\documentclass{article}
\usepackage{ijcai26}

\usepackage{times}
\usepackage{soul}
\usepackage{url}
\usepackage{makecell}
\usepackage[hidelinks]{hyperref}
\usepackage[utf8]{inputenc}
\usepackage[small]{caption}
\usepackage{graphicx}
\usepackage{amsmath, amssymb,amsfonts}
\usepackage{amsthm}
\usepackage{booktabs}
\usepackage{algorithm}
\usepackage{algorithmic}
\usepackage{subcaption}
\usepackage[switch]{lineno}
\usepackage{color}

\title{MemoVAD: Resource-Efficient Video Anomaly Detection via Dynamic Semantic Memory in Edge Computing Scenarios}

\author{
Guo Li$^1$
\and
Jiandian Zeng$^{1}$\thanks{\ Corresponding Author}\and
Yang Li$^2$\and
Zihao Peng$^1$\and
Ke Chen$^1$\And
Tian Wang$^3$\\
\affiliations
$^1$Institute of Artificial Intelligence and Future Networks, Beijing Normal University, Zhuhai, China\\
$^2$School of Computing and Artificial Intelligence, Southwest Jiaotong University, Chengdu, China\\
$^3$Engineering Research Center of Cloud-Edge Intelligent Collaboration on Big Data, Ministry of Education, Beijing Normal University, Zhuhai, Guangdong, China \\
\emails
\{liguo, pzh\_cs, kechen\}@mail.bnu.edu.cn,
\{jiandian, tianwang\}@bnu.edu.cn,
liyang23@swjtu.edu.cn
}

\begin{document}

\maketitle

\begin{abstract}
Deploying Video Anomaly Detection (VAD) in real-world surveillance faces a fundamental tension between the demand for high-level semantics to ensure effectiveness and the limited computational resources of edge devices. Vision–Language Models (VLMs) provide rich open-vocabulary semantics, but their latency and computational cost preclude on-device deployment. To address the challenge, we propose MemoVAD, an edge–cloud collaborative framework that selectively incorporates VLM semantics into streaming VAD. MemoVAD runs most inference on the edge with a lightweight detector and a causal Temporal Context Encoder (TCE) to model temporal dependencies. Specifically, we introduce an Uncertainty-Aware Gating (UAG) policy grounded in Subjective Logic to model perceived uncertainty and query the cloud-based VLM only for high-uncertainty and semantically novel clips. Besides, a Dynamic Semantic Memory (DSM) is designed to cache VLM-verified prototypes for efficient retrieval, enabling the edge model to progressively incorporate VLM-level semantics via a semantic adapter. Experiments on UCF-Crime and XD-Violence datasets via a real edge device show that MemoVAD substantially reduces communication overhead while surpassing state-of-the-art performance. The demo video is available at: \url{https://memovad2026.github.io/}.

\end{abstract}

\section{Introduction}
The widespread deployment of surveillance cameras in public spaces has resulted in an explosive growth of video data, creating an urgent demand for automated Video Anomaly Detection (VAD) \cite{shathik2025smart}. The primary objective of VAD is to identify abnormal events, such as accidents or criminal activities, within long sequences of normal activities. While recent advancements in deep learning have significantly improved detection accuracy, deploying the models in real-world scenarios remains a significant challenge. The difficulty arises primarily from the inherent conflict between the need for sophisticated semantic reasoning and the constrained computational resources available on edge devices \cite{ghasemi2024edgecloudai,guo2025edge}.

\begin{figure}[t]
    \centering
\includegraphics[width=0.9\linewidth]{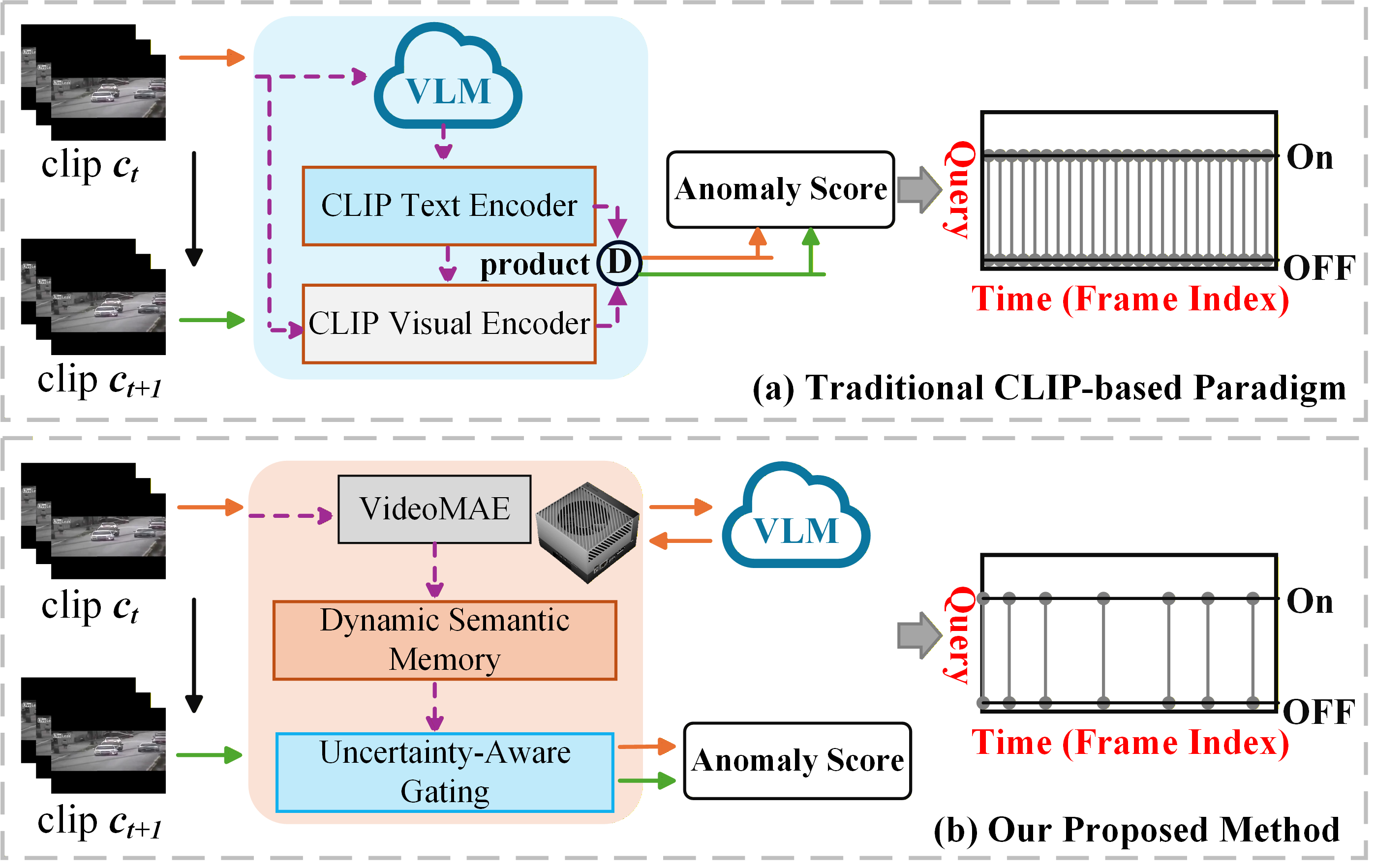}
    \caption{ Comparison of different WSVAD paradigms. (a) Traditional methods suffer from high communication overhead due to per-clip VLM querying. (b) Our proposed framework reduces costs by querying the VLM only for uncertain and novel samples.}
    \label{fig:motivation}
\end{figure}

Traditional VAD approaches typically rely on reconstruction or prediction paradigms that operate under unsupervised or weakly supervised settings \cite{wu2024vadclip,li2025anomize}. The methods generally learn the distribution of normal patterns and treat deviations as anomalies. Although computationally efficient enough for some edge applications, they often struggle to distinguish between genuine anomalies and benign distributional shifts, such as dynamic background changes or camera jitter \cite{sultani2018real,tian2021weakly}. Furthermore, the methods lack high-level semantic understanding, which restricts their ability to interpret complex scenes. Conversely, the emergence of Vision-Language Models (VLMs) has introduced a new paradigm where visual data is aligned with rich textual semantics through CLIP-based encoding \cite{deng2025vmad,li2025anomize}. VLMs demonstrate remarkable capability in zero-shot anomaly detection and semantic reasoning \cite{zanella2024harnessing,ye2025vera,shao2025eventvad}. However, their massive parameter counts and heavy computational requirements render them unsuitable for direct deployment on resource-constrained edge hardware.

To leverage the semantic power of VLMs without overwhelming edge resources, edge-cloud collaboration offers a promising direction \cite{zhang2025vavlm}. A naive solution is to transmit all video data to a cloud-based VLM for processing \cite{kumar2024cloud,wang2025holotrace,sharshar2025vision}. Nevertheless, it incurs huge bandwidth costs and high latency, which are unacceptable for real-time surveillance applications \cite{khan2022distributed,mondal2024toward,li2025e2ec}. Consequently, a critical research gap exists in developing a framework that can maintain the low latency of edge processing while selectively integrating the high-level semantic capabilities of cloud-resident large models.

To bridge the gap, we propose MemoVAD, a resource-efficient framework for video anomaly detection via dynamic semantic memory in edge computing scenarios. As shown in Fig.~\ref{fig:motivation}, anomalies typically exhibit recurring semantic patterns and manifest as temporally continuous segments. Therefore, continuous reliance on a computationally expensive VLM is unnecessary. Instead, we introduce a novel collaborative mechanism where a lightweight edge detector handles the majority of the inference workload and queries the remote VLM only when necessary. Specifically, MemoVAD introduces three components to balance semantic capability and edge efficiency. First, it adopts a lightweight detector that uses a frozen VideoMAE \cite{tong2022videomae} backbone for motion feature extraction, together with a causal Temporal Context Encoder (TCE) to model local temporal dependencies. Second, we propose an Uncertainty-Aware Gating (UAG) mechanism grounded in Subjective Logic \cite{sensoy2018evidential}, which estimates perceived uncertainty and triggers a VLM query when the edge model lacks sufficient evidence for a reliable decision. Finally, we develop a Dynamic Semantic Memory (DSM) that caches VLM-verified semantic prototypes on the edge. For semantically similar events, the model retrieves relevant knowledge from the memory instead of contacting the remote server, enabling efficient online knowledge distillation and progressively transferring VLM-level semantics to the edge model.

In summary, our main contributions are as follows: 
\begin{itemize} 
    \item We propose MemoVAD, a resource-efficient collaborative framework that harmonizes the speed of edge computing with the semantic reasoning of large VLMs. 
    \item We introduce an uncertainty-aware subjective-logic gate to query VLMs only under high perceived uncertainty, and a dynamic semantic memory caching VLM insights to reduce communication overhead while preserving accuracy.
    \item Our MemoVAD demonstrates remarkable improvements compared to various benchmarks across two public datasets, validating its superior performance.
\end{itemize}

\section{Related Works}

\subsection{ Weakly Supervised Video Anomaly Detection with VLMs} Due to the huge cost of frame-level annotations, current research predominantly focuses on Weakly Supervised Video Anomaly Detection (WSVAD), where only video-level labels are available \cite{mishra2024skeletal,abdalla2025video,liu2024generalized,wu2024weakly,karim2024real,zhang2023exploiting,pu2024learning}. Multiple Instance Learning (MIL) serves as the dominant framework in the domain, treating videos as bags of clips to distinguish anomalous bags from normal ones \cite{zhang2022dtfd,tang2023multiple,fang2024sam}. However, standard MIL-based detectors typically rely on discriminative feature embeddings that lack explicit semantic interpretability. To mitigate the issue, recent studies have integrated Vision-Language Models (VLMs), leveraging the zero-shot capabilities of models like CLIP \cite{radford2021learning} and GPT-4V to identify anomalies via textual prompting \cite{wu2024vadclip,ye2025vera,yang2024text,zanella2024harnessing,shao2025eventvad}. While VLMs demonstrate superior performance in recognizing semantically complex events, their substantial computational overhead poses a significant barrier to real-time deployment. Our work builds upon the efficient MIL formulation but enhances it by integrating explicit semantic knowledge. Specifically, to mitigate the computational cost of VLMs, we adopt an online knowledge distillation strategy that continuously updates the edge model via a dynamic memory mechanism.

\subsection{Efficient Edge-Cloud Collaboration} Deploying deep learning models on edge devices requires careful optimization to balance accuracy and efficiency \cite{guo2025edge,ghasemi2024edgecloudai}. Common techniques include model compression methods such as quantization, pruning, and lightweight architecture design. While the techniques reduce inference latency, they often degrade model capacity and performance. Edge-cloud collaborative intelligence seeks to mitigate the issue by offloading heavy computation to the cloud \cite{zhang2025vavlm,shathik2025smart}. Traditional offloading strategies decide which parts of a model to execute locally and which to transmit based on bandwidth and battery constraints. However, most existing collaborative frameworks focus on partition points within a fixed network architecture rather than dynamic interaction based on sample difficulty. Our approach differs by employing an uncertainty-driven policy that dynamically determines the necessity of cloud interaction. By utilizing Subjective Logic to quantify evidence sufficiency, MemoVAD ensures that communication resources are expended only on hard and novel samples, thereby achieving a superior balance between resource efficiency and detection performance.

\section{Methodology}

\begin{figure*}[t]
\centering
\includegraphics[width=0.95\textwidth]{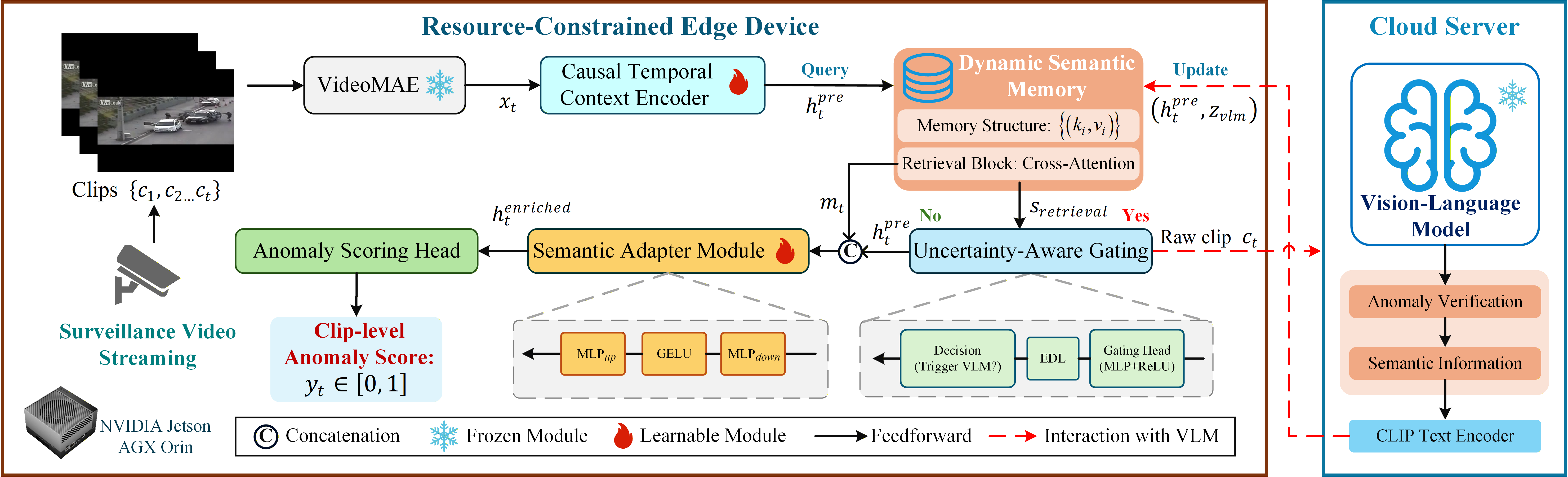}
\caption{The architecture of the MemoVAD system. }
\label{fig:architecture}
\end{figure*}

\subsection{Problem Definition}
\label{sec:problem definition}
We formulate WSVAD as a Multiple Instance Learning (MIL) task. Let $\mathcal{X} = \{(\mathcal{V}_i, Y_i)\}_{i=1}^{|\mathcal{X}|}$ denote the training set, where $\mathcal{V}_i$ is an untrimmed video and $Y_i \in \{0, 1\}$ is the video-level label. $Y_i=0$ indicates a normal video, while $Y_i=1$ implies the video contains at least one anomalous event. Each video $\mathcal{V}_i$ is divided into a sequence of $T$ non-overlapping clips $\left\{c_1, c_2, \ldots, c_T\right\}$.

The objective is to learn a clip-level anomaly scoring function $f_{\theta}(c_t) \rightarrow y_t \in [0, 1]$, such that $y_t$ is high for anomalous clips and low for normal ones. Under the MIL assumption, the relationship between clip scores and the video label is defined as:
\begin{equation}
Y_i = \max_{t=1}^{T} y_t.
\end{equation}

Specifically, for a normal bag $Y_i=0$, all clips are normal $y_t \approx 0$; for an anomalous bag $Y_i=1$, at least one clip is anomalous $y_t \approx 1$.


Unlike traditional centralized VAD, we consider an edge-cloud collaborative scenario. The local edge device has limited computational resources and must minimize communication with the remote VLM. The goal is to maximize detection performance while adhering to a strict communication budget:
\begin{equation}
\min \sum_{t=1}^{T} \mathbb{I}(\text{Query}_t) \quad \text{s.t. } \text{Accuracy} \geq \delta,
\end{equation}
where $\mathbb{I}(\cdot)$ is the indicator function for triggering a remote VLM query, and $\delta$ is the desired performance threshold.

\subsection{Overview}
We propose MemoVAD, a resource-efficient framework designed to bridge the gap between lightweight edge detection and semantic reasoning. As illustrated in Fig.~\ref{fig:architecture}, the framework operates under an edge-cloud collaborative paradigm to identify anomalies from the input video clips defined in Sec. \ref{sec:problem definition}. MemoVAD comprises three key components:
\begin{enumerate}
    \item \textbf{Edge-Resident Detector:} A lightweight network utilizing a frozen VideoMAE backbone for initial feature extraction and temporal modeling.
    \item \textbf{Uncertainty-Aware Gating (UAG):} A policy module that determines whether to query the VLM based on perceived uncertainty.
    \item \textbf{Dynamic Semantic Memory (DSM):} A continuously updating memory bank that stores VLM-verified prototypes, enabling the edge model to retrieve high-level semantic knowledge without recurring communication costs.
\end{enumerate}

\subsection{Edge-Resident Network}

\subsubsection{Feature Extraction}
To guarantee real-time inference on resource-constrained edge devices, we adopt VideoMAE-Small \cite{tong2022videomae} as our visual backbone. We freeze the pre-trained weights to preserve its generalizable motion features and prevent catastrophic forgetting during domain adaptation. Given an input video clip $c_t \in \mathbb{R}^{C \times F \times H \times W}$, the backbone extracts a compact feature vector $x_t$:

\begin{equation}
x_t=\mathcal{F}_{\text {backbone }}\left(c_t\right) \in \mathbb{R}^{D_{\text {stu }}},
\end{equation}
where $D_{\text{stu}}$ denotes the feature embedding dimension. For the input preprocessing, we resize frames to a standard spatial resolution of $H \times W$ and sample $F$ frames with a fixed temporal stride $\tau$.

\subsubsection{ Causal Temporal Context Encoder}
Since the backbone processes clips independently, the extracted feature $x_t$ lacks temporal context required for detecting complex anomalies. To bridge the gap, we introduce a lightweight Temporal Context Encoder (TCE) implemented as a 2-layer causal Transformer Encoder, which aggregates information from a fixed-length history window of $L$ clips. To retain sequential order information, learnable positional embeddings  are added to the input sequence before encoding:

\begin{equation}
h_t^{\text{pre}}=\operatorname{TCE}_{\text {causal }}\left(\left[x_{t-L+1}, \ldots, x_t\right]+P\right) ,
\label{eq:tce}
\end{equation}
where $P$ denotes learnable positional embeddings. The output $h_t^{\text{pre}}\in\mathbb{R}^{D_{\text{stu}}}$ is the pre-enrichment contextual student feature for clip $t$. In deployment, we implement the causal Transformer with KV-cache for streaming inference, enabling amortized constant-time updates per incoming clip.

\subsection{Dynamic Semantic Memory (DSM)}
DSM is designed to realize \emph{online knowledge distillation} by caching VLM reasoning results and enabling semantic retrieval on the edge.

\subsubsection{Memory Structure}
We structure the memory as a set of key-value pairs $\mathcal{M}=\{(k_i,v_i)\}_{i=1}^{N}$ with memory size $N$.
\begin{itemize}
    \item \textbf{Key} $k_i \in \mathbb{R}^{D_{\text{stu}}}$ stores the pre-enrichment student feature $h^{\text{pre}}$ for a VLM-confirmed event. It ensures that keys share the same distribution as incoming queries.
    \item \textbf{Value} $v_i \in \mathbb{R}^{D_{\text{vlm}}}$ stores the corresponding semantic embedding produced by the teacher VLM, serving as high-level knowledge to be distilled.
\end{itemize}

\subsubsection{Memory Retrieval}
For each incoming clip feature $h_t^{\text{pre}}$, we retrieve relevant semantic information from $\mathcal{M}$ using multi-head cross-attention. The query is derived from the student feature, while the memory keys and values serve as $K$ and $V$:
\begin{equation}
Q=h_t^{\text{pre}} W_Q,\quad K=\mathcal{M}_{\text{key}} W_K,\quad V=\mathcal{M}_{\text{val}} W_V.
\end{equation}

The retrieved semantic information $m_t$ is computed via the attention mechanism:
\begin{equation}
m_t=\operatorname{Attention}(Q, K, V)=\operatorname{Softmax}\left(\frac{Q K^T}{\sqrt{d_k}}\right) V.
\end{equation}

Simultaneously, to quantify retrieval confidence for gating, we compute the maximum cosine similarity between $h_t^{\text{pre}}$ and stored keys:
\begin{equation}
s_{\text{retrieval}} = \max_{i \in \{1, \dots, N\}} \left( \frac{h_t^{\text{pre}} \cdot k_i}{\|h_t^{\text{pre}}\|_2 \|k_i\|_2} \right)\in[-1,1].
\end{equation}

The metric $s_{\text{retrieval}}$ measures how well the current input matches the known anomalous prototypes in the memory. 

\subsubsection{Semantic Adapter Module (SAM)}
Directly fusing semantic prototypes into the student space may cause distribution mismatch. We introduce a lightweight Semantic Adapter Module (SAM) to align and fuse $m_t$ into the student feature space.

We first project the retrieved semantic prototype to the student dimension via a lightweight projection $\phi:\mathbb{R}^{D_{\text{vlm}}}\rightarrow\mathbb{R}^{D_{\text{stu}}}$ (e.g., a single linear layer). Then we concatenate and pass through a bottleneck adapter:
\begin{equation}
\Delta h_t = \operatorname{MLP}_{up}\left(\sigma\left(\operatorname{MLP}_{down}\left(\operatorname{Concat}\left[h_t^{\text{pre}}, \phi(m_t)\right]\right)\right)\right),
\end{equation}
where $\operatorname{MLP}_{down}$ compresses by ratio $r=4$, $\sigma$ is GELU, and $\operatorname{MLP}_{up}$ restores to $D_{\text{stu}}$. The enriched student feature is obtained by a gated residual connection:
\begin{equation}
h_t^{\text{enriched}} = \operatorname{LayerNorm}\left(h_t^{\text{pre}} + \alpha \cdot \Delta h_t\right),
\end{equation}
where $\alpha$ is a learnable scalar initialized to $0$ to prevent negative transfer in early training.

\subsection{Uncertainty-Aware Gating (UAG)}
Continuous VLM querying is extremely expensive. UAG requests VLM assistance only for \emph{hard or novel} samples, using perceived uncertainty and memory confidence.

\subsubsection{Gating Policy}

Softmax confidence cannot reliably separate aleatoric and perceived uncertainty. We adopt Subjective Logic~\cite{sensoy2018evidential} to estimate perceived uncertainty via evidential learning.

Given the pre-enrichment feature $h_t^{\text{pre}}$, the gating head predicts non-negative evidence:
\begin{equation}
\mathbf{e}_t=\operatorname{ReLU}\left(\operatorname{MLP}\left(h_t^{\text{pre}}\right)\right),\quad \mathbf{e}_t=\left[e_{t,0},e_{t,1}\right],
\end{equation}
for the Normal and Anomalous classes. Evidence defines a Dirichlet distribution $\mathrm{Dir} (\boldsymbol{\alpha}_t)$:

\begin{equation}
\boldsymbol{\alpha}_t=\mathbf{e}_t+\mathbf{1}, \quad S_t=\sum_k \alpha_{t, k} ,
\end{equation}
where $S_t$ represents evidence strength. The perceived uncertainty is defined as inverse total evidence:
\begin{equation}
u_t=\frac{K}{S_t}=\frac{2}{\alpha_{t,0}+\alpha_{t,1}},\quad (K=2).
\end{equation}

To improve robustness under domain shift where models can be over-confident yet wrong, we trigger VLM queries when the case is uncertain and novel. The triggering condition is:

\begin{equation}
\operatorname{Trigger}_t=\mathbb{I}\left(u_t>\tau_{\text {unc }} \wedge s_{\text {retrieval }}<\tau_{\text {sim }}\right) ,
\label{eq:gate policy}
\end{equation}
where $\tau_{\text{unc}},\tau_{\text{sim}}$ are hyper-parameters trading off accuracy and communication cost.

\subsubsection{VLM Inference and Memory Update}

When triggered, the raw clip $c_t$ is transmitted to the remote VLM. We uniformly sample $F$ frames from $c_t$ and provide a lightweight prompt for anomaly verification. The VLM outputs:
(1) Anomaly Verification, deciding whether the clip truly contains an anomaly; and
(2) Semantic Extraction, producing a text-aligned embedding $z_{\text{vlm}}\in\mathbb{R}^{D_{\text{vlm}}}$.

If the VLM confirms an anomaly, we write the new knowledge into memory:
\begin{equation}
\mathcal{M} \leftarrow \mathcal{M} \cup \left\{\left(h_t^{\text{pre}}, z_{\text{vlm}}\right)\right\}.
\end{equation}

\paragraph{Online Semantic-Diversity Replacement.}
To keep a fixed memory budget $N_{\max}$ while avoiding quadratic recomputation, we maintain a redundancy score for each key:
\begin{equation}
\rho_i=\max_{j\neq i}\cos(k_i,k_j).
\end{equation}
When inserting a new key $k_{\text{new}}$, we compute $\cos(k_{\text{new}},k_i)$ for all stored keys and update affected $\rho_i$ incrementally. If the memory is full, we remove the most redundant key $\arg\max_i \rho_i$, yielding an online replacement with amortized $O(ND_{\text{stu}})$ per insertion.

\subsection{Objective Function}
\label{sec:training_objectives}

The model is trained with weak video-level labels. Let $S(\cdot)$ denote the anomaly classifier head followed by a Sigmoid. The predicted anomaly score for the $t$-th clip is:
\begin{equation}
y_t = S\left(h_t^{\text{enriched}}\right)\in[0,1].
\end{equation}

\subsubsection{MIL Ranking Loss}
We formulate weakly supervised VAD as a Multiple-Instance Learning task. Normal videos are negative bags $\mathcal{B}_n$ and anomalous videos are positive bags $\mathcal{B}_a$. Using the standard MIL assumption, the video-level score is represented by the maximum clip score. The ranking loss is:
\begin{equation}
\mathcal{L}_{\text{mil}}=\max\left(0, m - \max_{t \in \mathcal{B}_a} y_t + \max_{t \in \mathcal{B}_n} y_t \right),
\end{equation}
\label{eq:mil}
where $m\in(0,1)$ is the margin hyperparameter.

\subsubsection{Semantic Distillation Loss}
To distill semantic knowledge from the VLM, we minimize the distance between student and teacher representations. We introduce a projection head $\psi:\mathbb{R}^{D_{\text{stu}}}\rightarrow\mathbb{R}^{D_{\text{vlm}}}$ (two-layer MLP) and apply supervision only when teacher signals are available. The distillation loss is:

\begin{equation}
\begin{aligned}
\mathcal{L}_{\text {distill }} = \frac{1}{\sum_{t=1}^{T} \mathbb{M}_t + \epsilon} \sum_{t=1}^{T} \mathbb{M}_t \cdot \| \psi\left(h_t^{\text {enriched }}\right) - \\
\text{stop\_grad}\left(z_{vlm}\right) \|_2^2 ,
\end{aligned}
\end{equation}
where $\epsilon$ is a small constant for numerical stability. Here, $\mathbb{M}_t=1$ if the VLM is triggered for clip $t$, and $0$ otherwise. The stop-gradient operator treats the VLM as a fixed semantic anchor.

\subsubsection{Temporal Smoothness Loss}
We impose temporal continuity and sparsity priors:
\begin{equation}
\mathcal{L}_{\text{smooth}}=\sum_{t=1}^{T-1}\left(y_t-y_{t+1}\right)^2+\lambda_{sp}\sum_{t=1}^{T}y_t,
\label{eq:smooth}
\end{equation}
where the first term enforces smoothness, and the second encourages sparsity of anomaly predictions. $\lambda_{sp}$ is a sparsity hyper-parameter that balances the two constraints. It prevents the trivial solution where the model predicts high scores for all clips.


\subsubsection{Gating Head Supervision}
We supervise the evidential gating head following Evidential Deep Learning~\cite{sensoy2018evidential}. Let $Y_{\text{bag}}\in\{0,1\}$ be the video-level label. To reduce label noise under MIL, we apply the loss only to the top-scoring clip(s) in each bag:
\begin{equation}
\mathcal{L}_{\text{gate}} = \mathcal{L}_{\text{EDL}}(\boldsymbol{\alpha}_t, Y_{\text{bag}}).
\end{equation}

\subsubsection{Total Loss}
The final objective function is a weighted sum of the components:

\begin{equation}
\mathcal{L}_{\text {total }} = \mathcal{L}_{\text {mil }} + \lambda_1 \mathcal{L}_{\text {distill }} + \lambda_2 \mathcal{L}_{\text {smooth }} + \lambda_3 \mathcal{L}_{\text {gate }} ,
\label{eq:loss}
\end{equation}
where $\lambda_1, \lambda_2,$ and $\lambda_3$ are pre-defined hyperparameters that balance the trade-off between anomaly detection, semantic alignment, and uncertainty estimation.

\section{Experiments}
\subsection{Experimental Settings} 
\noindent\textbf{Datasets.} We evaluate MemoVAD on two large-scale weakly supervised video anomaly detection benchmarks. \textbf{(1) UCF-Crime} \cite{sultani2018real} contains 1,900 real-world surveillance videos spanning 13 anomaly categories and normal activities.
\textbf{(2) XD-Violence} \cite{wu2020not} consists of 4,754 videos collected from movies and games, providing audio-visual signals. In our experiments, we use only the visual modality. 


\noindent\textbf{Metrics.} Following standard protocols, we employ the Area Under the Receiver Operating Characteristic Curve (\textbf{AUC}) for UCF-Crime and Average Precision (\textbf{AP}) for XD-Violence to evaluate detection performance. In addition to detection performance, we strictly evaluate resource efficiency in edge scenarios. Specifically, we report Throughput (\textbf{FPS}) and Latency (\textbf{s}) to verify real-time capabilities, and define Communication Rate (\textbf{CR}) as the percentage of clips that trigger a VLM query. 

\noindent\textbf{Baselines.} For comparison, Sultani et al.~\cite{sultani2018real}, RTFM~\cite{tian2021weakly}, CRFD~\cite{wu2021learning}, MSL~\cite{li2022self}, MGFN~\cite{chen2023mgfn}, UR-DMU~\cite{zhou2023dual}, CLIP-TSA~\cite{joo2023clip}, TPWNG~\cite{yang2024text}, VadCLIP~\cite{wu2024vadclip}, OVVAD~\cite{li2025anomize}, and EventVAD~\cite{shao2025eventvad} are chosen as baselines.

\begin{table}[t]
\centering
\setlength{\tabcolsep}{2pt}  
\resizebox{\linewidth}{!}{
\begin{tabular}{lcccc} 
\toprule 
Method & Ref. & Backbone & AUC (\%)$\uparrow$ & AP (\%)$\uparrow$ \\
\midrule 
Sultani et al. & CVPR'18  & C3D & 77.92 & 73.20  \\
RTFM  & ICCV'21  & I3D & 84.30 & 77.81  \\
CRFD   & TIP'21   & I3D & 84.89 & 75.90  \\
MSL        & AAAI'22  & I3D & 85.62 & 78.58  \\
MGFN     & AAAI'23  & I3D & 86.67 & 80.11  \\
UR-DMU  & AAAI'23  & I3D & 86.97 & 81.66  \\
CLIP-TSA  & ICIP'23  & CLIP & 87.58 & 82.17  \\
TPWNG    & CVPR'24  & CLIP & 87.79 & 83.68  \\
VadCLIP   & AAAI'24  & CLIP & 88.02 & 84.51  \\
OVVAD    & CVPR'25  & CLIP & 86.40 & 69.31  \\
EventVAD     & MM'25    & CLIP & 87.51 & 64.04  \\ 
\midrule
\textbf{MemoVAD (Ours)} & - & VideoMAE-S & \textbf{89.45} & \textbf{85.97}  \\
\bottomrule 
\end{tabular}}
\caption{Comparison results on UCF-Crime (AUC) and XD-Violence (AP). Best results are highlighted in \textbf{bold}.}
\label{tab:Compasion}
\end{table}

\begin{table}[t]
\centering
\small
\setlength{\tabcolsep}{2pt} 
\begin{tabular}{lcccc} 
\toprule
Method & FPS $\uparrow$ & Latency (s)$\downarrow$ & CR$_{\text{UCF}} $(\%)$\downarrow$ & CR$_{\text{XD}} $(\%)$\downarrow$ \\
\midrule
I3D baseline & 23.7 & 0.868 & 0.00 & 0.00 \\
CLIP-based baseline & 11.5 & 1.512 & 100.0 & 100.0 \\
\midrule
\textbf{MemoVAD (Ours)} & \textbf{36.2} & \textbf{0.475} & \textbf{8.63} & \textbf{15.72} \\
\bottomrule
\end{tabular}
\caption{Efficiency on Jetson AGX Orin (streaming, batch=1). FPS and Latency are measured end-to-end. CR: Percentage of clips querying the remote VLM.}
\label{tab:efficiency}
\end{table}

\subsection{Comparison with State-of-the-Art Methods}

\noindent\textbf{Main Results.} As presented in Table~\ref{tab:Compasion}, MemoVAD achieves 89.45\% AUC on UCF-Crime and 85.97\% AP on XD-Violence, consistently outperforming prior approaches with C3D/I3D or CLIP-based feature backbones. Such consistent gains across two widely used benchmarks indicate the strong effectiveness and robustness of our method under different anomaly categories and evaluation metrics. In particular, MemoVAD improves the best baselines TPWNG and VadCLIP by an absolute margin of 1.66\% and 1.43\% AUC on UCF-Crime, and 2.29\% and 1.46\% AP on XD-Violence, respectively. Notably, the gains are achieved by utilizing the frozen VideoMAE-S as a resource-efficient student backbone, demonstrating that accurate anomaly detection does not strictly require computationally intensive foundation models.

\begin{table}[t]
\centering
\small
\setlength{\tabcolsep}{2pt} 
\begin{tabular}{cccc@{\hskip 8pt}cccc} 
\toprule
TCE & DSM & UAG & SAM & AUC (\%)$\uparrow$ & FPS $\uparrow$ & Latency (s)$\downarrow$ & CR (\%)$\downarrow$ \\
\midrule
$\times$   & $\times$   & $\times$   & $\times$   & 72.33 & 40.6 & 0.266 & 0.00 \\
\checkmark & $\times$   & $\times$   & $\times$   & 79.56 & 39.8 & 0.353 & 0.00 \\
\checkmark & \checkmark & $\times$   & $\times$   & 90.10 & 12.8 & 1.481 & 100.0 \\
\checkmark & \checkmark & \checkmark & $\times$   & 87.85 & 37.5 & 0.462 & 8.63 \\
\checkmark & \checkmark & \checkmark & \checkmark & \textbf{89.45} & \textbf{36.2} & \textbf{0.475} & \textbf{8.63} \\
\bottomrule
\end{tabular}
\caption{Ablation study of key components on UCF-Crime. $\checkmark$ and $\times$ denote the inclusion and exclusion of each module, respectively.}
\label{tab:ablation}
\end{table}

\begin{table}[t]
\centering
\small
\setlength{\tabcolsep}{2.5pt}
\begin{tabular}{lc@{\hskip 8pt}cc}
\toprule
Gating Strategy & Metric & AUC (\%)$\uparrow$ & CR (\%)$\downarrow$ \\
\midrule
Softmax Entropy & Confidence & 86.12 & 12.45 \\
Evidential Uncertainty (Only) & Uncertainty & 87.05 & 9.80 \\
Retrieval Similarity (Only) & Similarity & 88.20 & 14.20 \\
\midrule
\textbf{Hybrid (Ours)} & Unc. + Sim. & \textbf{89.45} & \textbf{8.63} \\
\bottomrule
\end{tabular}
\caption{Comparison of different gating policies on UCF-Crime.}
\label{tab:gating_ablation}
\end{table}

\begin{table}[t]
\centering
\setlength{\tabcolsep}{7pt} 
\begin{tabular}{lcc} 
\toprule
Update Policy & AUC (\%)$\uparrow$ & AP (\%)$\uparrow$ \\
\midrule
First-In-First-Out (FIFO) & 87.34 & 83.10 \\
Random Replacement & 86.90 & 82.55 \\
Least Recently Used (LRU) & 88.10 & 84.20 \\
\midrule
\textbf{Semantic-Diversity (Ours)} & \textbf{89.45} & \textbf{85.97} \\
\bottomrule
\end{tabular}
\caption{Ablation on memory update policies. AUC and AP are selected as metrics.}
\label{tab:memory_policy}
\end{table}

\begin{figure}[t] 
    \centering
    \begin{subfigure}{0.48\linewidth}
        \centering
        \includegraphics[width=\linewidth]{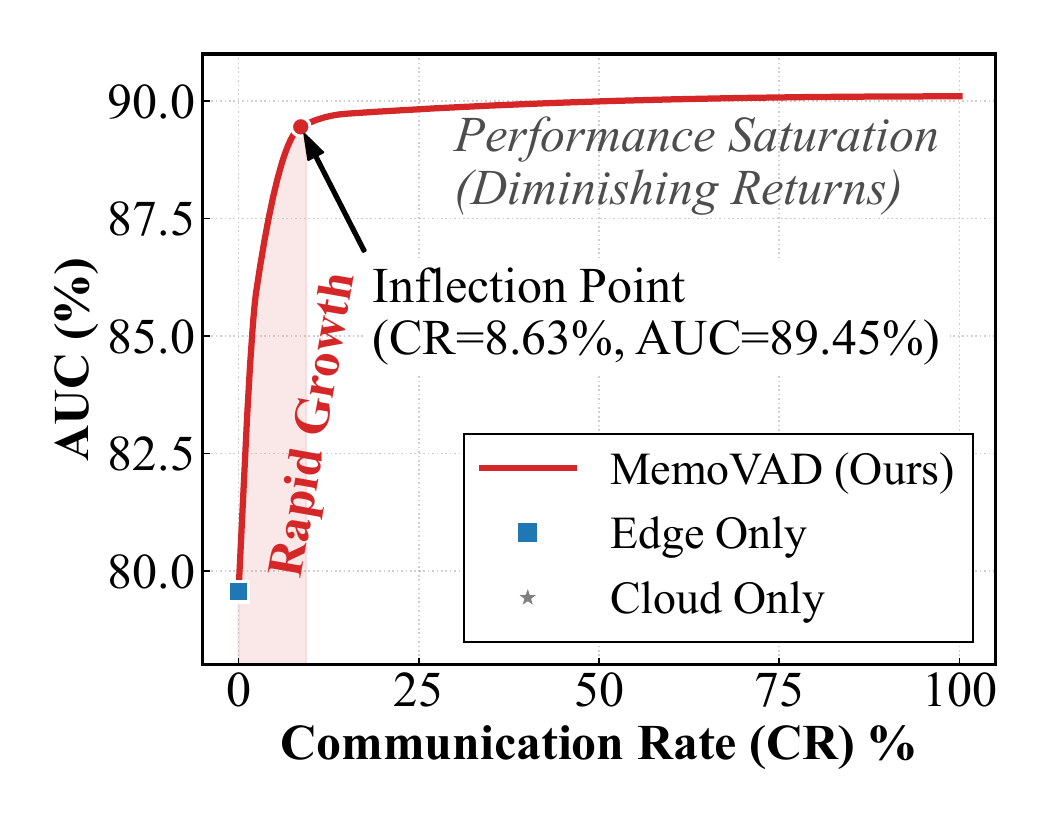} 
        \caption{ } 
        \label{fig:tradeoff_a}
    \end{subfigure}
    \hfill 
    \begin{subfigure}{0.48\linewidth}
        \centering
        \includegraphics[width=\linewidth]{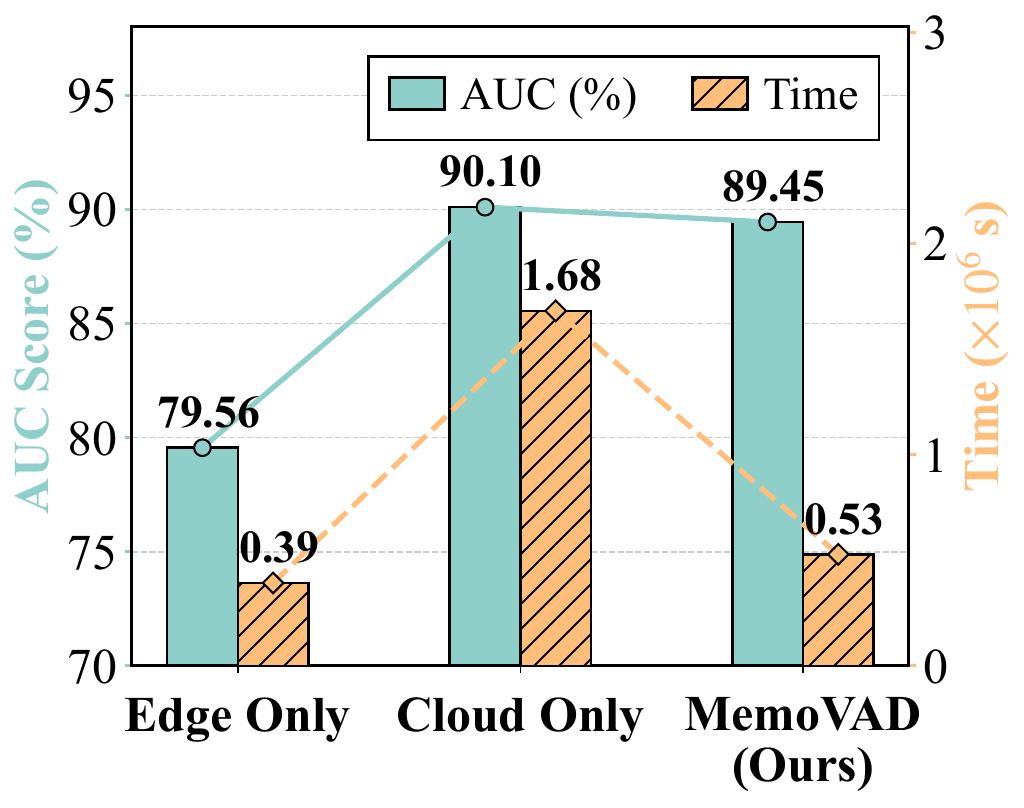}
        \caption{}
        \label{fig:tradeoff_b}
    \end{subfigure}
    \caption{Trade-off evaluation on UCF-Crime. (a) AUC vs. Communication Rate. (b) AUC vs. Time}
    \label{fig:tradeoff}
\end{figure}

\begin{figure*}[t] 
    \centering
    \begin{subfigure}{0.24\textwidth}
        \centering
        \includegraphics[width=\linewidth]{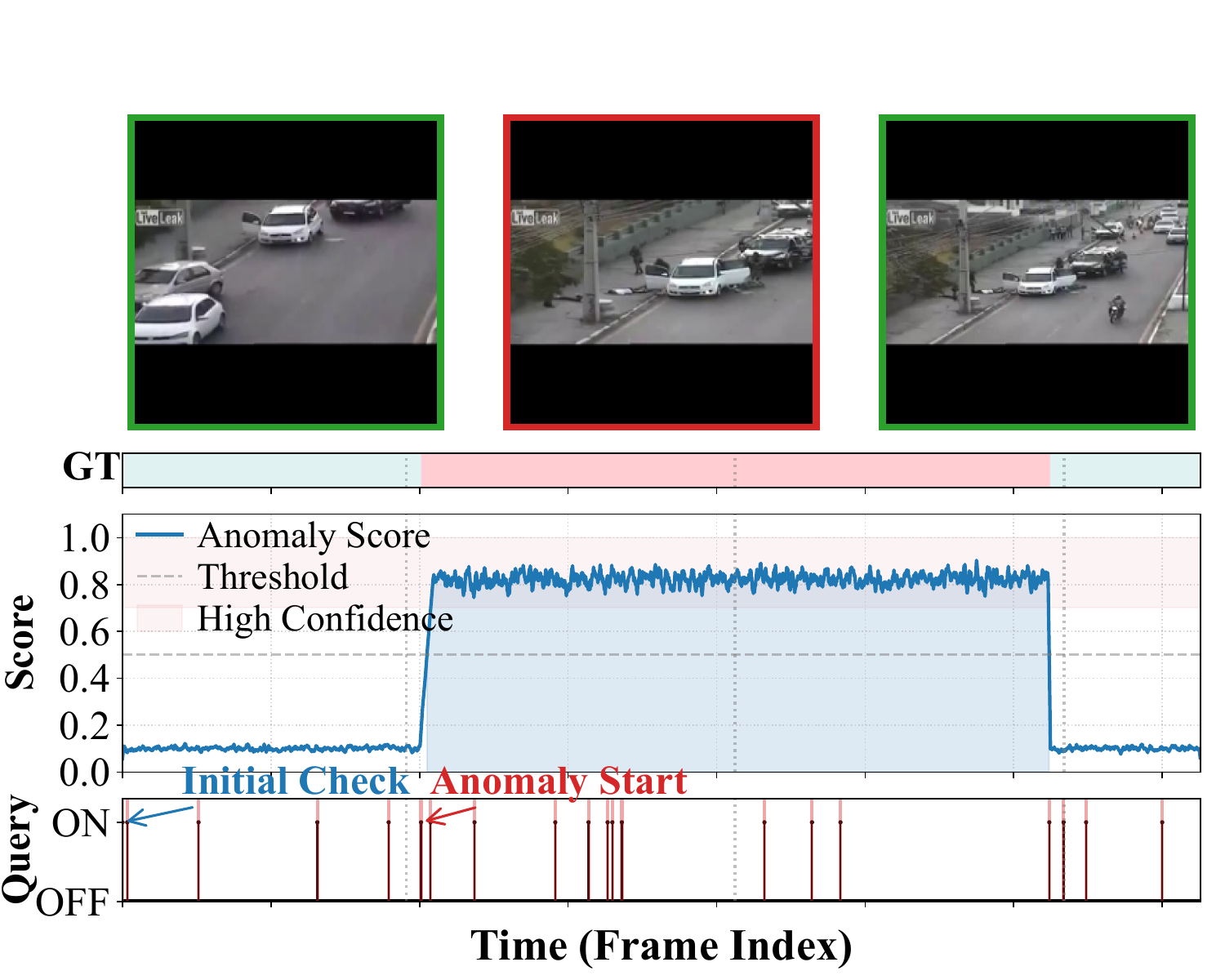} 
        \caption{Arrest} 
        \label{fig:Arrest}
    \end{subfigure}
    \hfill 
    \begin{subfigure}{0.24\textwidth}
        \centering
        \includegraphics[width=\linewidth]{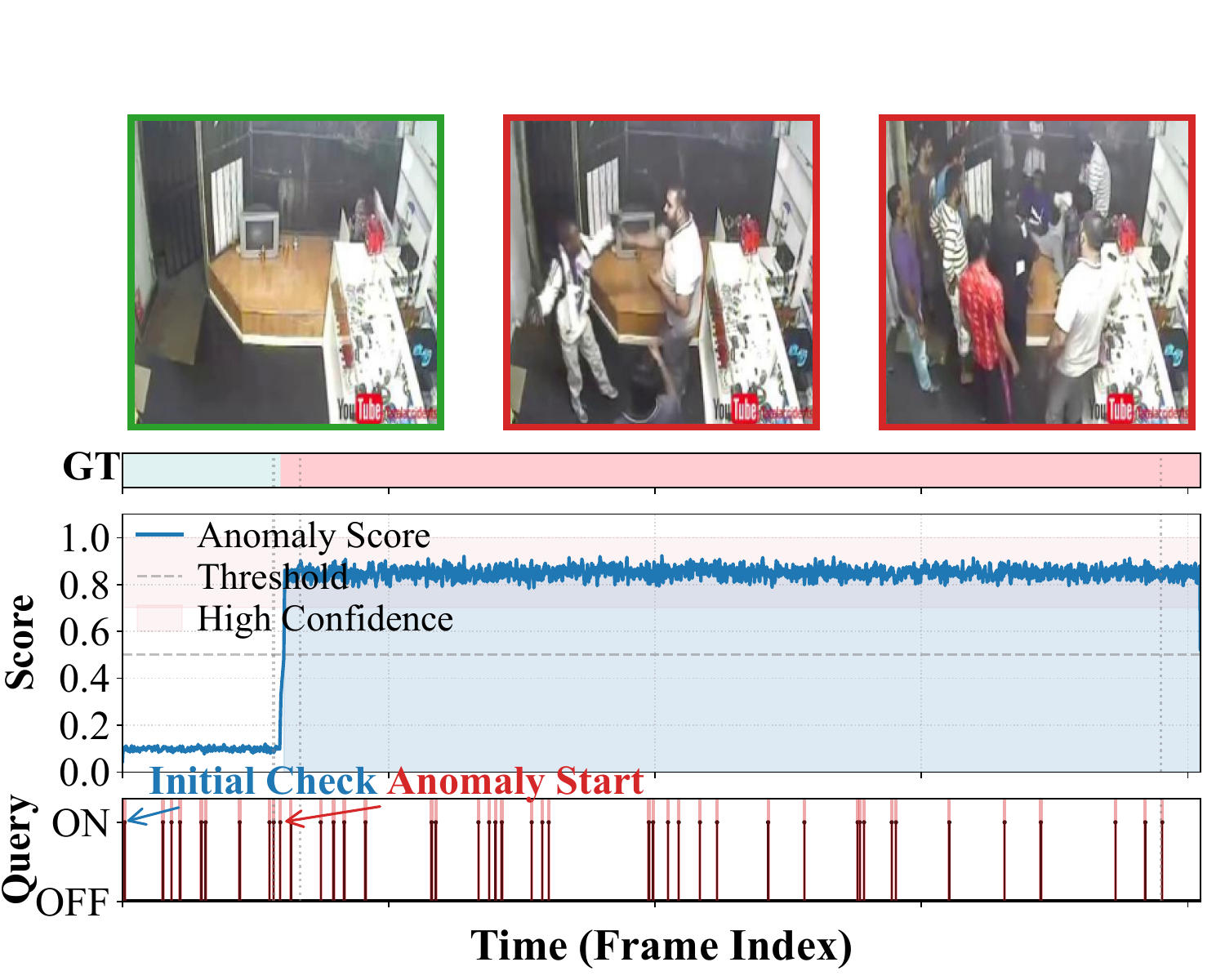}
        \caption{Assault}
        \label{fig:Assault}
    \end{subfigure}
    \hfill
    \begin{subfigure}{0.24\textwidth}
        \centering
        \includegraphics[width=\linewidth]{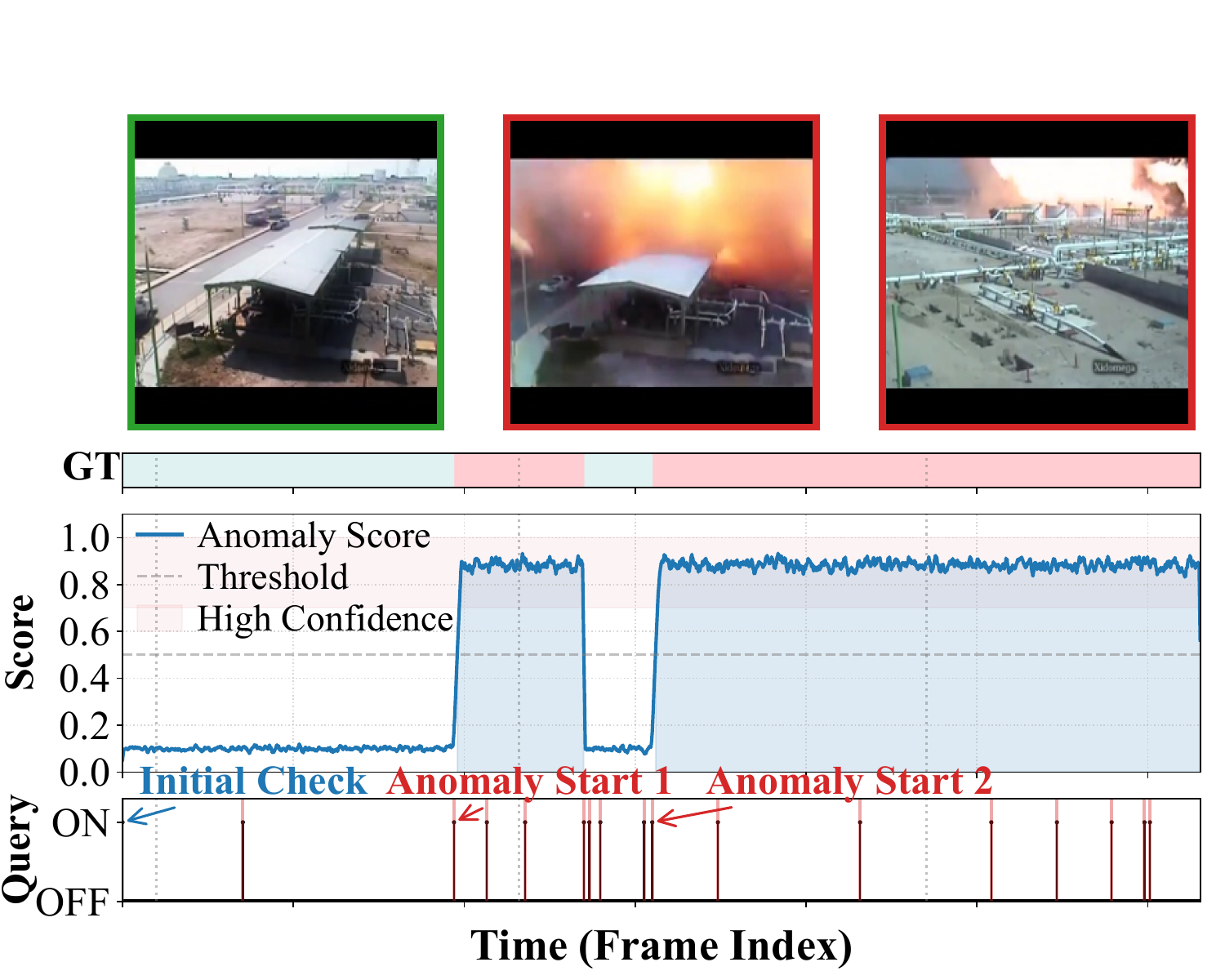}
        \caption{Explosion}
        \label{fig:Explosion}
    \end{subfigure}
    \hfill
    \begin{subfigure}{0.24\textwidth}
        \centering
        \includegraphics[width=\linewidth]{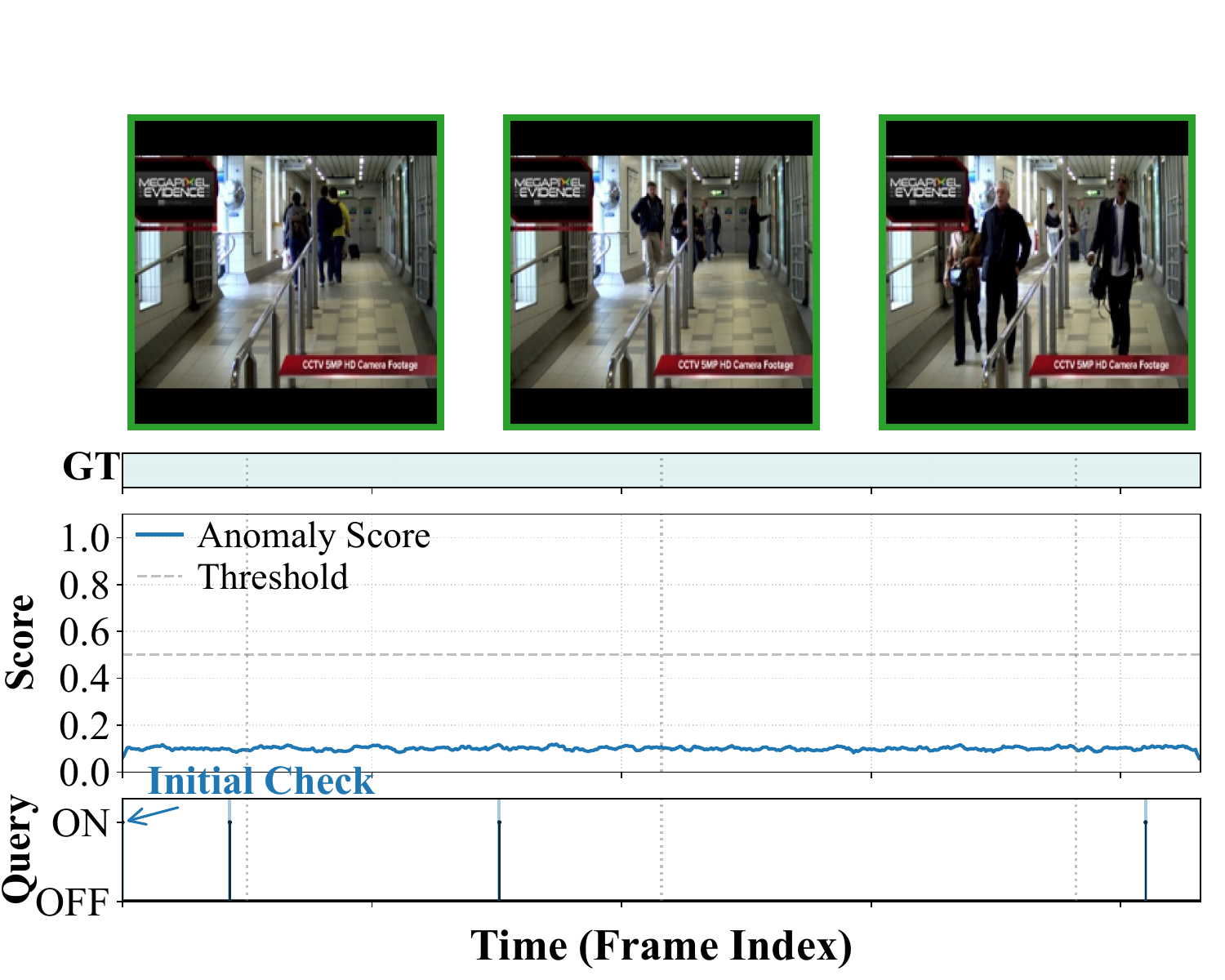}
        \caption{Normal Video}
        \label{fig:Normal_Video}
    \end{subfigure}
    \caption{Qualitative visualization of MemoVAD on abnormal and normal scenarios. The rows from top to bottom illustrate: input frames with detection status (Green: Normal, Red: Anomaly), Ground Truth (GT), predicted Anomaly Score, and the VLM Query Trigger signal. }
    \label{fig:qualitative}
\end{figure*}

\noindent\textbf{Efficiency Results.} MemoVAD is explicitly designed for real-time anomaly detection under an edge-collaborative setting, where both on-device computation and communication costs are critical. As detailed in Table~\ref{tab:efficiency}, MemoVAD maintains real-time capability on the Jetson AGX Orin by achieving a throughput of 36.2 FPS. The performance represents a substantial improvement over the I3D baseline, which operates exclusively on the edge at 23.7 FPS. Furthermore, MemoVAD significantly outpaces the CLIP-based baseline that relies continuously on remote VLM support, running at only 11.5 FPS. The throughput improvement is also reflected in the latency metrics, where MemoVAD achieves a rapid response time of 0.475 seconds, markedly outperforming both the CLIP-based baseline and the I3D-based baseline. Such a performance gain validates the efficacy of utilizing the lightweight VideoMAE-S architecture, which avoids the computational bottlenecks often associated with the heavier semantic backbones employed by competing methods. Beyond local computational efficiency, MemoVAD significantly minimizes the dependency on remote VLM access. While the CLIP-based baseline necessitates a 100\% query rate to function, our method initiates VLM queries for merely 8.63\% and 15.72\% of the clips on the UCF-Crime and XD-Violence datasets respectively. By processing the vast majority of clips locally, MemoVAD effectively circumvents the bandwidth limitations and latency penalties often associated with expensive remote inference.

\subsection{Ablation Studies}

\paragraph{Effectiveness of Key Components.}
Table \ref{tab:ablation} summarizes the incremental contributions of each module. Starting with the baseline in Row 1, the frozen VideoMAE backbone achieves 72.33\% AUC at 40.6 FPS, but is constrained by the lack of temporal modeling. Incorporating the causal TCE as shown in Row 2 effectively aggregates historical motion cues and boosts AUC to 79.56\% with a negligible latency overhead resulting in 39.8 FPS. To gauge the maximum potential of semantic reasoning, we evaluate a teacher forcing variant with DSM in Row 3 that queries the VLM for every clip. While the configuration yields a peak AUC of 90.10\%, it incurs huge costs including a 100\% communication rate and a throughput drop to 12.8 FPS with a high latency of 1.481 seconds, which renders it impractical for edge deployment. To mitigate the issue, the UAG mechanism in Row 4 selectively triggers queries only for clips with high-uncertainty and novelty, reducing communication to 8.63\% and restoring a real-time throughput of 37.5 FPS with 0.462 seconds, while incurring only a modest AUC drop to 87.85\%. Finally, the introduced SAM  alleviates the feature distribution mismatch between the student and teacher. By aligning retrieved semantics, it recovers the AUC to 89.45\% and closely matches the upper bound while maintaining a high efficiency of 36.2 FPS and a marginal latency of 0.475 seconds.

\paragraph{Analysis of Gating Policies.} To validate the design of our Uncertainty-Aware Gating, we compare the proposed method against standard baselines as detailed in Table \ref{tab:gating_ablation}. The baseline utilizing solely Softmax confidence yields suboptimal performance of 86.12\%, primarily due to its inability to discriminate between hard samples and out-of-distribution data. While employing evidential uncertainty alone improves precision, it fails to capture novel semantic events that are statistically confident yet semantically unfamiliar. Consequently, our hybrid approach achieves the optimal synergy by triggering VLM queries exclusively when the model exhibits perceived uncertainty or when the input remains semantically distinct from existing memory prototypes.

\paragraph{Effectiveness of Memory Dynamics.} We further investigate the impact of memory management on long-term learning stability in Table \ref{tab:memory_policy}. Naive strategies, such as FIFO or Random replacement, result in the catastrophic forgetting of rare anomaly prototypes, leading to a performance degradation exceeding 2\%. In contrast, our Online Semantic-Diversity Replacement ensures the retention of a spanning set of diverse abnormal information, thereby maximizing the utility of the fixed storage budget of 2,048 slots.

\paragraph{Efficiency-Accuracy Trade-off.} Fig.~\ref{fig:tradeoff} illustrates the balance between computational efficiency and detection accuracy. To characterize the operational envelope of MemoVAD, we systematically modulate the gating hyperparameters, specifically the uncertainty threshold $\tau_{\text{unc}}$ and the similarity threshold $\tau_{\text{sim}}$. Fig.~\ref{fig:tradeoff_a} reveals a smooth Pareto frontier where the proposed method maintains an AUC exceeding 89\% even when the communication overhead is constrained to less than 10\%. 
Furthermore, we provide a comparative analysis against the baseline deployment paradigms in Fig.~\ref{fig:tradeoff_b}. Specifically, while exhibiting a marginal degradation in AUC compared to the Cloud-only approach, our method significantly reduces the inference latency. Notably, the total runtime of MemoVAD is approximately one-third of that required by the Cloud-only paradigm, demonstrating its practicality for real-time edge computing scenarios. Such performance significantly surpasses the baselines and underscores the adaptability of our framework to fluctuating network bandwidths in real-world deployments.


\subsection{Qualitative Analysis}
\paragraph{Qualitative Visualization.} Fig.~\ref{fig:qualitative} details the inference dynamics of MemoVAD across diverse scenarios. The blue curves denote the predicted anomaly scores and the black vertical stems indicate VLM query triggers. 
As presented in Fig.~\ref{fig:Arrest}, the query frequency spikes immediately at the anomaly onset to resolve high perceived uncertainty. Upon the termination of the event, the system rapidly validates the return to a normal state, resulting in a rapid decay of the anomaly score. Similarly, in Fig.~\ref{fig:Assault}, where the anomaly persists until the end of the video, the system maintains robust recognition throughout the abnormal event duration.
Notably, in Fig.~\ref{fig:Explosion}, the MemoVAD successfully validates the interval between events to confirm the temporary restoration of normality, demonstrating its ability to capture complex temporal dependencies. In the normal video depicted in Fig.~\ref{fig:Normal_Video}, our system maintains consistently low anomaly scores with extremely sparse updates, effectively minimizing computational costs in the absence of semantic shifts. Overall, MemoVAD precisely localizes events matching ground-truth red borders, while meeting rate limits via content-adaptive resources.

\begin{figure}[t] 
    \centering
    \begin{subfigure}{0.48\linewidth}
        \centering
        \includegraphics[width=\linewidth]{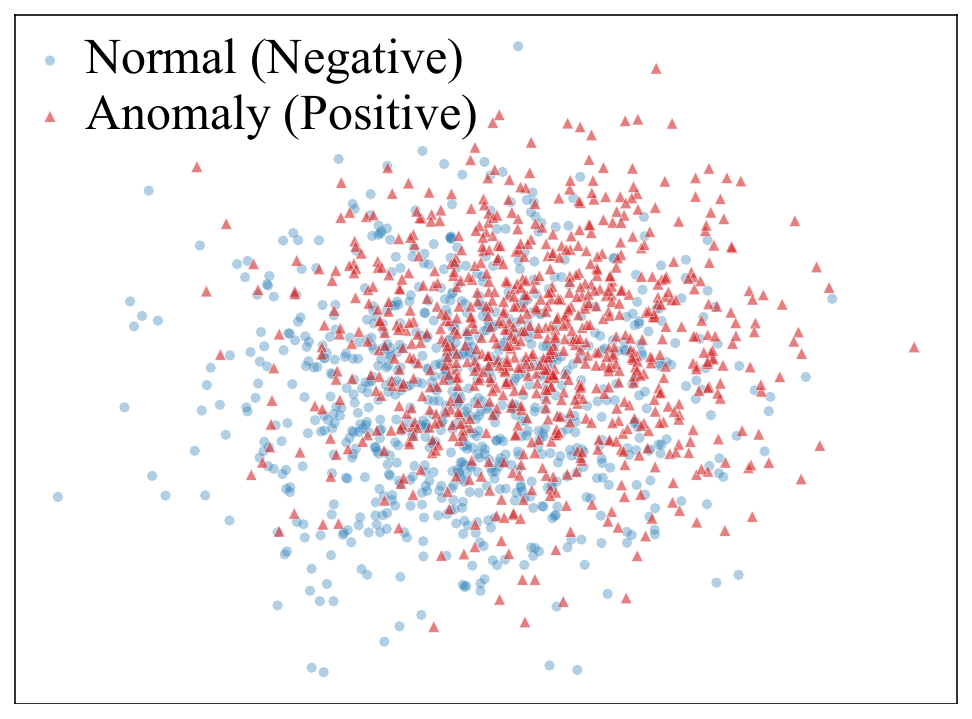} 
        \caption{Pre-enrichment Features} 
        \label{fig:tsne_a}
    \end{subfigure}
    \hfill 
    \begin{subfigure}{0.48\linewidth}
        \centering
        \includegraphics[width=\linewidth]{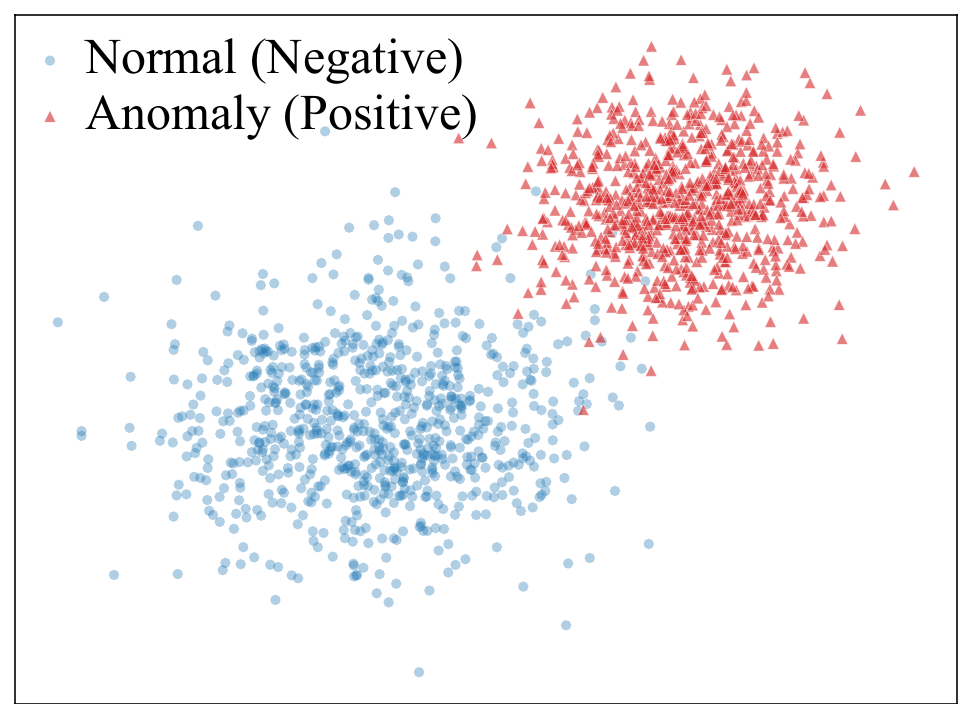}
        \caption{Enriched Features}
        \label{fig:tsne_b}
    \end{subfigure}
    \caption{Feature visualization via t-SNE on UCF-Crime dataset. }
    \label{fig:tsne}
\end{figure}

\paragraph{Feature Discrimination Visualization.} To intuitively evaluate the quality of learned representations, we visualize the feature distributions on the UCF-Crime dataset using t-SNE, as depicted in Fig.~\ref{fig:tsne}. The visualization in Fig.~\ref{fig:tsne_a} reveals that the baseline VideoMAE exhibits a high degree of feature entanglement, where anomalous samples exhibit heavily overlap with normal patterns due to the lack of discriminative semantic guidance. In contrast, as illustrated in Fig.~\ref{fig:tsne_b}, benefiting from specific optimization and semantic enrichment of MemoVAD, the learned features form distinct and compact clusters with clear decision boundaries. It further demonstrates that MemoVAD effectively disentangles anomalous features away from the normal distribution.

\section{Conclusion}

In this work, we proposed MemoVAD, a framework that strategically bridges a lightweight edge detector with a powerful remote VLM. Instead of continuously relying on the cloud, our system learns to seek remote guidance only when uncertain, caching the insights locally to improve over time. Our experiments confirm that the uncertainty-driven collaboration yields state-of-the-art performance with minimal bandwidth usage, demonstrating a practical path for deploying foundation model capabilities in real-world surveillance without being overwhelmed by their computational weight.

Future work will explore extending the paradigm to collaborative edge computing scenarios, where distributed devices can share semantic knowledge and jointly maintain memory. We will also investigate long-term adaptation under non-stationary environments, including scene shifts and evolving anomaly patterns. In addition, more robust memory update and validation mechanisms will be studied to reduce the influence of noisy VLM feedback while preserving the efficiency benefits of selective remote querying.

\appendix



\section*{Acknowledgments}
The above work was supported in part by the Joint Funds of the National Natural Science Foundation of China under Grant U25A20436, Guangxi Key Research \& Development Program (FN2504240036, 2025FN96441087), the National Natural Science Foundation of China (NSFC) (62372047, 62302049), Guangdong S\&T Programme (No. 2025B0101120006), the Natural Science Foundation of Guangdong Province (2024A1515011323), the Supplemental Funds for Major Scientific Research Projects of Beijing Normal University, Zhuhai (ZHPT2023002), the Fundamental Research Funds for the Central Universities, Guangdong Province Educational Science Planning Research Project under 2025JKZG069, Higher Education Research Topics of Guangdong Association of Higher Education in the 14th Five-Year Plan under 24GYB207, Natural Science Foundation of Sichuan under Grant 2026NSFSC1459 and support from the Interdisciplinary Intelligence Super Computer Center of Beijing Normal University at Zhuhai.

\bibliographystyle{named}
\bibliography{ijcai26}

@article{shathik2025smart,
  title={SMART VISION SYSTEMS FOR PUBLIC SAFETY: REAL-TIME CROWD MONITORING AND ANOMALY DETECTION IN URBAN SPACES USING DEEP LEARNING AND EDGE COMPUTING},
  author={Shathik, J Anvar},
  journal={International Journal of Applied Mathematics},
  volume={38},
  number={6s},
  pages={720--743},
  year={2025}
}

@inproceedings{ghasemi2024edgecloudai,
  title={EdgeCloudAI: Edge-Cloud Distributed Video Analytics},
  author={Ghasemi, Mahshid and Kostic, Zoran and Ghaderi, Javad and Zussman, Gil},
  booktitle={Proceedings of the 30th Annual International Conference on Mobile Computing and Networking},
  pages={1778--1780},
  year={2024}
}

@article{guo2025edge,
  title={Edge-Cloud Collaborative Real-Time Video Object Detection for Industrial Surveillance Systems},
  author={Guo, Siyan and Zhao, Cong and Yang, Shusen and Liang, Yingying and Wang, Yimeng and Han, Qing},
  journal={IEEE Intelligent Systems},
  year={2025},
  publisher={IEEE}
}

@article{deng2025vmad,
  title={Vmad: Visual-enhanced multimodal large language model for zero-shot anomaly detection},
  author={Deng, Huilin and Luo, Hongchen and Zhai, Wei and Guo, Yanming and Cao, Yang and Kang, Yu},
  journal={IEEE Transactions on Automation Science and Engineering},
  year={2025},
  publisher={IEEE}
}

@inproceedings{ye2025vera,
  title={Vera: Explainable video anomaly detection via verbalized learning of vision-language models},
  author={Ye, Muchao and Liu, Weiyang and He, Pan},
  booktitle={Proceedings of the Computer Vision and Pattern Recognition Conference},
  pages={8679--8688},
  year={2025}
}

@inproceedings{wu2024vadclip,
  title={Vadclip: Adapting vision-language models for weakly supervised video anomaly detection},
  author={Wu, Peng and Zhou, Xuerong and Pang, Guansong and Zhou, Lingru and Yan, Qingsen and Wang, Peng and Zhang, Yanning},
  booktitle={Proceedings of the AAAI Conference on Artificial Intelligence},
  volume={38},
  number={6},
  pages={6074--6082},
  year={2024}
}

@inproceedings{shao2025eventvad,
  title={Eventvad: Training-free event-aware video anomaly detection},
  author={Shao, Yihua and He, Haojin and Li, Sijie and Chen, Siyu and Long, Xinwei and Zeng, Fanhu and Fan, Yuxuan and Zhang, Muyang and Yan, Ziyang and Ma, Ao and others},
  booktitle={Proceedings of the 33rd ACM International Conference on Multimedia},
  pages={2586--2595},
  year={2025}
}

@inproceedings{zanella2024harnessing,
  title={Harnessing large language models for training-free video anomaly detection},
  author={Zanella, Luca and Menapace, Willi and Mancini, Massimiliano and Wang, Yiming and Ricci, Elisa},
  booktitle={Proceedings of the IEEE/CVF Conference on Computer Vision and Pattern Recognition},
  pages={18527--18536},
  year={2024}
}

@inproceedings{li2025anomize,
  title={Anomize: Better Open Vocabulary Video Anomaly Detection},
  author={Li, Fei and Liu, Wenxuan and Chen, Jingjing and Zhang, Ruixu and Wang, Yuran and Zhong, Xian and Wang, Zheng},
  booktitle={Proceedings of the Computer Vision and Pattern Recognition Conference},
  pages={29203--29212},
  year={2025}
}

@article{zhang2025vavlm,
  title={VaVLM: Toward Efficient Edge-Cloud Video Analytics With Vision-Language Models},
  author={Zhang, Yang and Wang, Hanling and Bai, Qing and Liang, Haifeng and Zhu, Peican and Muntean, Gabriel-Miro and Li, Qing},
  journal={IEEE Transactions on Broadcasting},
  year={2025},
  publisher={IEEE}
}

@inproceedings{wang2025holotrace,
  title={HoloTrace: LLM-based Bidirectional Causal Knowledge Graph for Edge-Cloud Video Anomaly Detection},
  author={Wang, Hanling and Li, Qing and Chen, Li and Kang, Haidong and Ma, Fei and Jiang, Yong},
  booktitle={Proceedings of the 33rd ACM International Conference on Multimedia},
  pages={6510--6519},
  year={2025}
}

@article{sharshar2025vision,
  title={Vision-language models for edge networks: A comprehensive survey},
  author={Sharshar, Ahmed and Khan, Latif U and Ullah, Waseem and Guizani, Mohsen},
  journal={IEEE Internet of Things Journal},
  year={2025},
  publisher={IEEE}
}

@article{li2025e2ec,
  title={E2EC: Edge-to-Edge Collaboration for Efficient Real-Time Video Surveillance Inference},
  author={Li, Guo and Zeng, Jiandian and Peng, Zihao and Liang, Yuzhu and Zheng, Xi and Wang, Tian},
  journal={IEEE Transactions on Mobile Computing},
  year={2025},
  publisher={IEEE}
}

@article{mondal2024toward,
  title={Toward energy-efficient and cost-effective task offloading in mobile edge computing for intelligent surveillance systems},
  author={Mondal, Manash Kumar and Banerjee, Sourav and Das, Debashis and Ghosh, Uttam and Al-Numay, Mohammed S and Biswas, Utpal},
  journal={IEEE Transactions on Consumer Electronics},
  volume={70},
  number={1},
  pages={4087--4094},
  year={2024},
  publisher={IEEE}
}

@article{khan2022distributed,
  title={Distributed inference in resource-constrained IoT for real-time video surveillance},
  author={Khan, Muhammad Asif and Hamila, Ridha and Erbad, Aiman and Gabbouj, Moncef},
  journal={IEEE Systems Journal},
  volume={17},
  number={1},
  pages={1512--1523},
  year={2022},
  publisher={IEEE}
}

@article{kumar2024cloud,
  title={Cloud-based video streaming services: Trends, challenges, and opportunities},
  author={Kumar, Tajinder and Sharma, Purushottam and Tanwar, Jaswinder and Alsghier, Hisham and Bhushan, Shashi and Alhumyani, Hesham and Sharma, Vivek and Alutaibi, Ahmed I},
  journal={CAAI Transactions on Intelligence Technology},
  volume={9},
  number={2},
  pages={265--285},
  year={2024},
  publisher={Wiley Online Library}
}

@article{abdalla2025video,
  title={Video anomaly detection in 10 years: A survey and outlook},
  author={Abdalla, Moshira and Javed, Sajid and Al Radi, Muaz and Ulhaq, Anwaar and Werghi, Naoufel},
  journal={Neural Computing and Applications},
  pages={1--44},
  year={2025},
  publisher={Springer}
}

@article{mishra2024skeletal,
  title={Skeletal video anomaly detection using deep learning: Survey, challenges, and future directions},
  author={Mishra, Pratik K and Mihailidis, Alex and Khan, Shehroz S},
  journal={IEEE Transactions on Emerging Topics in Computational Intelligence},
  volume={8},
  number={2},
  pages={1073--1085},
  year={2024},
  publisher={IEEE}
}

@article{liu2024generalized,
  title={Generalized video anomaly event detection: Systematic taxonomy and comparison of deep models},
  author={Liu, Yang and Yang, Dingkang and Wang, Yan and Liu, Jing and Liu, Jun and Boukerche, Azzedine and Sun, Peng and Song, Liang},
  journal={ACM Computing Surveys},
  volume={56},
  number={7},
  pages={1--38},
  year={2024},
  publisher={ACM New York, NY}
}

@inproceedings{radford2021learning,
  title={Learning transferable visual models from natural language supervision},
  author={Radford, Alec and Kim, Jong Wook and Hallacy, Chris and Ramesh, Aditya and Goh, Gabriel and Agarwal, Sandhini and Sastry, Girish and Askell, Amanda and Mishkin, Pamela and Clark, Jack and others},
  booktitle={International Conference on Machine Learning},
  pages={8748--8763},
  year={2021},
  organization={PmLR}
}

@inproceedings{wu2020not,
  title={Not only look, but also listen: Learning multimodal violence detection under weak supervision},
  author={Wu, Peng and Liu, Jing and Shi, Yujia and Sun, Yujia and Shao, Fangtao and Wu, Zhaoyang and Yang, Zhiwei},
  booktitle={European Conference on Computer Vision},
  pages={322--339},
  year={2020},
  organization={Springer}
}

@inproceedings{wu2024weakly,
  title={Weakly supervised video anomaly detection and localization with spatio-temporal prompts},
  author={Wu, Peng and Zhou, Xuerong and Pang, Guansong and Yang, Zhiwei and Yan, Qingsen and Wang, Peng and Zhang, Yanning},
  booktitle={Proceedings of the 32nd ACM International Conference on Multimedia},
  pages={9301--9310},
  year={2024}
}

@inproceedings{karim2024real,
  title={Real-time weakly supervised video anomaly detection},
  author={Karim, Hamza and Doshi, Keval and Yilmaz, Yasin},
  booktitle={Proceedings of the IEEE/CVF Winter Conference on Applications of Computer Vision},
  pages={6848--6856},
  year={2024}
}

@inproceedings{zhang2023exploiting,
  title={Exploiting completeness and uncertainty of pseudo labels for weakly supervised video anomaly detection},
  author={Zhang, Chen and Li, Guorong and Qi, Yuankai and Wang, Shuhui and Qing, Laiyun and Huang, Qingming and Yang, Ming-Hsuan},
  booktitle={Proceedings of the IEEE/CVF Conference on Computer Vision and Pattern Recognition},
  pages={16271--16280},
  year={2023}
}

@article{pu2024learning,
  title={Learning prompt-enhanced context features for weakly-supervised video anomaly detection},
  author={Pu, Yujiang and Wu, Xiaoyu and Yang, Lulu and Wang, Shengjin},
  journal={IEEE Transactions on Image Processing},
  year={2024},
  publisher={IEEE}
}

@inproceedings{zhang2022dtfd,
  title={Dtfd-mil: Double-tier feature distillation multiple instance learning for histopathology whole slide image classification},
  author={Zhang, Hongrun and Meng, Yanda and Zhao, Yitian and Qiao, Yihong and Yang, Xiaoyun and Coupland, Sarah E and Zheng, Yalin},
  booktitle={Proceedings of the IEEE/CVF Conference on Computer Vision and Pattern recognition},
  pages={18802--18812},
  year={2022}
}

@inproceedings{tang2023multiple,
  title={Multiple instance learning framework with masked hard instance mining for whole slide image classification},
  author={Tang, Wenhao and Huang, Sheng and Zhang, Xiaoxian and Zhou, Fengtao and Zhang, Yi and Liu, Bo},
  booktitle={Proceedings of the IEEE/CVF International Conference on Computer Vision},
  pages={4078--4087},
  year={2023}
}

@inproceedings{fang2024sam,
  title={Sam-mil: A spatial contextual aware multiple instance learning approach for whole slide image classification},
  author={Fang, Heng and Huang, Sheng and Tang, Wenhao and Huangfu, Luwen and Liu, Bo},
  booktitle={Proceedings of the 32nd ACM International Conference on Multimedia},
  pages={6083--6092},
  year={2024}
}

@article{sensoy2018evidential,
  title={Evidential deep learning to quantify classification uncertainty},
  author={Sensoy, Murat and Kaplan, Lance and Kandemir, Melih},
  journal={Advances in Neural Information Processing Systems},
  volume={31},
  year={2018}
}

@inproceedings{tian2021weakly,
  title={Weakly-supervised video anomaly detection with robust temporal feature magnitude learning},
  author={Tian, Yu and Pang, Guansong and Chen, Yuanhong and Singh, Rajvinder and Verjans, Johan W and Carneiro, Gustavo},
  booktitle={Proceedings of the IEEE/CVF International Conference on Computer Vision},
  pages={4975--4986},
  year={2021}
}

@inproceedings{sultani2018real,
  title={Real-world anomaly detection in surveillance videos},
  author={Sultani, Waqas and Chen, Chen and Shah, Mubarak},
  booktitle={Proceedings of the IEEE Conference on Computer Vision and Pattern Recognition},
  pages={6479--6488},
  year={2018}
}

@inproceedings{li2022self,
  title={Self-training multi-sequence learning with transformer for weakly supervised video anomaly detection},
  author={Li, Shuo and Liu, Fang and Jiao, Licheng},
  booktitle={Proceedings of the AAAI Conference on Artificial Intelligence},
  volume={36},
  number={2},
  pages={1395--1403},
  year={2022}
}

@inproceedings{yang2024text,
  title={Text prompt with normality guidance for weakly supervised video anomaly detection},
  author={Yang, Zhiwei and Liu, Jing and Wu, Peng},
  booktitle={Proceedings of the IEEE/CVF Conference on Computer Vision and Pattern Recognition},
  pages={18899--18908},
  year={2024}
}

@article{wu2021learning,
  title={Learning causal temporal relation and feature discrimination for anomaly detection},
  author={Wu, Peng and Liu, Jing},
  journal={IEEE Transactions on Image Processing},
  volume={30},
  pages={3513--3527},
  year={2021},
  publisher={IEEE}
}

@inproceedings{chen2023mgfn,
  title={Mgfn: Magnitude-contrastive glance-and-focus network for weakly-supervised video anomaly detection},
  author={Chen, Yingxian and Liu, Zhengzhe and Zhang, Baoheng and Fok, Wilton and Qi, Xiaojuan and Wu, Yik-Chung},
  booktitle={Proceedings of the AAAI Conference on Artificial Intelligence},
  volume={37},
  number={1},
  pages={387--395},
  year={2023}
}

@inproceedings{zhou2023dual,
  title={Dual memory units with uncertainty regulation for weakly supervised video anomaly detection},
  author={Zhou, Hang and Yu, Junqing and Yang, Wei},
  booktitle={Proceedings of the AAAI Conference on Artificial Intelligence},
  volume={37},
  number={3},
  pages={3769--3777},
  year={2023}
}

@inproceedings{joo2023clip,
  title={Clip-tsa: Clip-assisted temporal self-attention for weakly-supervised video anomaly detection},
  author={Joo, Hyekang Kevin and Vo, Khoa and Yamazaki, Kashu and Le, Ngan},
  booktitle={2023 IEEE International Conference on Image Processing (ICIP)},
  pages={3230--3234},
  year={2023},
  organization={IEEE}
}

@article{tong2022videomae,
  title={Videomae: Masked autoencoders are data-efficient learners for self-supervised video pre-training},
  author={Tong, Zhan and Song, Yibing and Wang, Jue and Wang, Limin},
  journal={Advances in Neural Information Processing Systems},
  volume={35},
  pages={10078--10093},
  year={2022}
}

\end{document}